\documentclass[10pt,twocolumn,letterpaper]{article}

\usepackage{cvpr}
\usepackage{graphicx}
\usepackage{multirow}
\usepackage{comment}
\usepackage[T1]{fontenc}
\usepackage[accsupp]{axessibility}
\usepackage[table]{xcolor}

\definecolor{cvprblue}{rgb}{0.21,0.49,0.74}
\usepackage[pagebackref,breaklinks,colorlinks,allcolors=cvprblue]{hyperref}

\def\paperID{63}
\def\confName{CVPR}
\def\confYear{2026}

\usepackage{xspace}

\title{Seeing without Looking: Do Vision-Language Benchmarks Really Test Vision?}

\author{
Zixuan Lan\thanks{Equal contribution.} \textsuperscript{\hspace{0.2em}}\thanks{Work done as a research intern at Stony Brook University.}\\
University of Chicago\\
{\tt\small zixuanlan@uchicago.edu}
\and
\hspace{1.5cm}Luzhe Sun\footnotemark[1]\\
\hspace{1.5cm}Toyota Technological Institute at Chicago\\
{\tt\small \hspace{1.5cm}luzhesun@ttic.edu}
\and
\hspace{-2.1cm}Matthew R. Walter\\
\hspace{-2.1cm}Toyota Technological Institute at Chicago\\
{\tt\small \hspace{-2.1cm}mwalter@ttic.edu}
\and
\hspace{0.9cm}Jiawei Zhou\\
\hspace{0.9cm}Stony Brook University\\
{\tt\small \hspace{0.9cm}jiawei.zhou.1@stonybrook.edu}
}

\begin{document}
\maketitle
\begin{abstract}
Benchmark accuracy is often implicitly assumed to reflect grounded visual understanding in vision--language models (VLMs), yet it remains unclear to what extent such scores truly reflect reliance on visual evidence. Motivated by a surprising observation that removing a substantial fraction of image tokens only degrades model performance very slightly on a widely used hallucination benchmark, we systematically investigate this mismatch in a set of open-source VLMs. Our analysis spans multiple levels of granularity, spanning global visual degradation, localized occlusion, question reformulation, answer-space expansion, and decision-level analyses beyond standard accuracy. We further complement these behavioral results with a layer-wise analysis of vision-token geometry. Throughout the experiments, we find that although VLMs do incorporate visual input, their predictions are less sensitive to the loss of fine-grained visual evidence that standard accuracy should have suggested. Even when the final prediction remains unchanged, the model's internal support for the correct answer may already be weakened. We further complement a representation level analysis, which shows increasing similarity among visual tokens in deeper layers, providing a possible explanation for our findings. Together, these results suggest that current benchmarks are not sufficient to reliably evaluate fine-grained visual grounding in VLMs.
\end{abstract}
\section{Introduction}
\label{sec:intro}

\begin{figure}[t]
    \centering
    \includegraphics[width=0.98\linewidth]{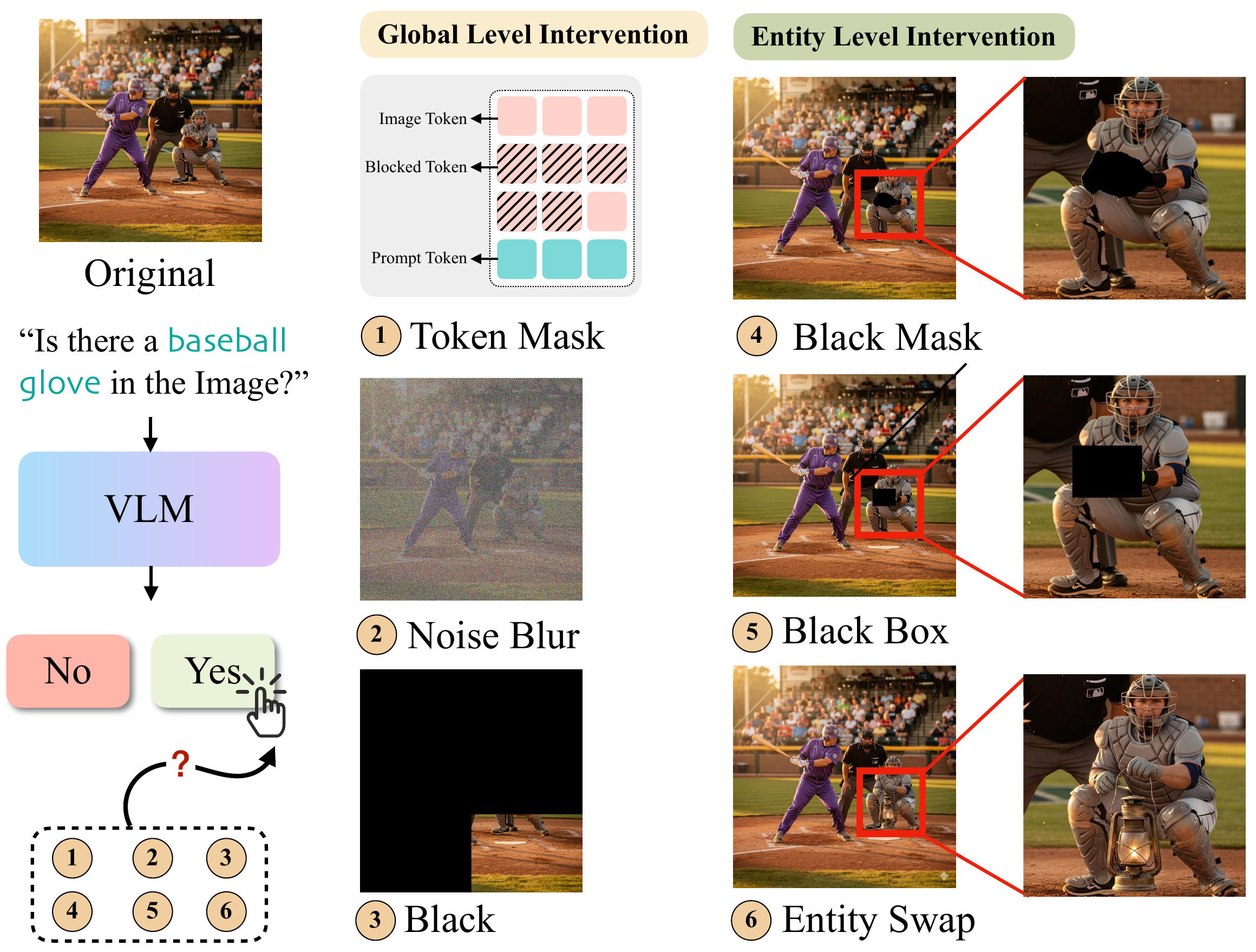}
    \caption{We intervene on the input image at both global and entity levels while keeping the question fixed. Global-level interventions weaken overall visual evidence through token masking, noise blur, and black, whereas entity-level interventions manipulate the question-relevant region via black mask, black box, and entity swap. In this example, despite substantial degradation or alteration of the visual evidence, the model's final answer can remain unchanged and still predict that a baseball glove is present.}
    \label{fig:teaser}
\end{figure}

Vision--language models (VLMs) have achieved strong performance on a growing collection of benchmarks intended to evaluate grounded visual understanding~\cite{fu2025mmecomprehensiveevaluationbenchmark,huang2024surveyevaluationmultimodallarge, kogilathota2026halp}. Yet a fundamental question remains unresolved: to what extent do these benchmark scores truly reflect a model's reliance on visual evidence, rather than its ability to exploit coarse visual cues, redundant representations, or strong language priors? This question is increasingly important, as benchmark results are often taken as evidence of progress in visual reasoning and hallucination mitigation~\cite{fu2025mmecomprehensiveevaluationbenchmark,song2025head, huang2024surveyevaluationmultimodallarge, chenhalc, fang2025enhancing}.

In this work, we investigate this question through a simple yet revealing observation. On POPE~\cite{li2023evaluatingobjecthallucinationlarge}, a widely used benchmark for probing object hallucination \cite{kogilathota2026halp}, we find that randomly removing a substantial fraction of image tokens often leads to little degradation in overall accuracy. At face value, this result is surprising: if benchmark success truly depends on robust visual grounding, one would expect performance to decline as visual evidence is removed. More importantly, however, this phenomenon is inherently ambiguous. It may reflect (i) redundancy in visual representations, (ii) the possibility that the benchmark can be solved using only coarse-grained visual information, or (iii) the presence of strong language-side regularities that allow models to maintain correct answers even when visual input is weakened~\cite{lee-etal-2025-vlind,tong2024eyeswideshutexploring}. This observation points to a deeper measurement problem: benchmark accuracy, the dominant validation signal for a growing range of VLM methods (including work on data selection and training improvement \cite{ding2020statistical, song2025head}, hallucination mitigation \cite{chenhalc}, model robustness and safety \cite{jiang2025robustifying}, and inference-time efficiency \cite{li-etal-2025-text}, such as reducing or dropping image tokens \cite{fang2025enhancing}) may not be an effective metric for assessing fine-grained visual grounding.

If accuracy can remain high with weakened visual evidence, as in Figure \ref{fig:teaser}, then improvements (or robustness) on these benchmarks may not reliably indicate improved visual grounding, and can lead to overconfident conclusions about what a method actually preserves or improves.

Motivated by this puzzle, we conduct a comprehensive analysis of how VLM benchmark performance changes under progressively weakened, localized, or semantically altered visual evidence. Our study spans multiple benchmarks and multiple levels of intervention, including token-level reduction of visual input, global image degradation, removal of localized entity evidence, semantic manipulation of queried entities, and analyses of decision uncertainty under alternative answer spaces. Instead of treating benchmark accuracy as a sufficient summary of model capability, we examine when it remains stable, when it fails, and what kinds of visual dependence such behavior actually reflects.

Overall, our results suggest that while VLMs do rely on visual input, benchmark accuracy often overstates the degree of fine-grained visual grounding. Strong performance can remain intact despite the loss of detailed visual evidence, supported by coarse visual cues, redundancy, evaluation constraints, and language priors~\cite{lee-etal-2025-vlind,tong2024eyeswideshutexploring}. This gap between accuracy and grounding should be regarded as a core issue in VLM evaluation, motivating more diagnostic ways of assessing model capability.

Our contributions are three-fold:

\begin{itemize}
    \item We identify a  phenomenon in VLM evaluation: standard benchmark metrics often do not even degrade much as expected when corresponding visual evidence is substantially weakened or removed.
    \item We develop a detailed multi-level diagnostic framework to probe what benchmark performance actually depends on.
    \item We show that current VLM benchmarks can substantially overestimate fine-grained visual grounding, revealing a systematic gap between benchmark success and genuine reliance on question-relevant visual evidence.
\end{itemize}
\section{Background}

\paragraph{Hallucination and grounding benchmarks.}
Recent work has introduced a range of benchmarks for evaluating hallucination and grounding in vision--language models \cite{li2023evaluatingobjecthallucinationlarge, kaul2025throneobjectbasedhallucinationbenchmark, fu2025mmecomprehensiveevaluationbenchmark, sadana2025iso}. POPE \cite{li2023evaluatingobjecthallucinationlarge} formulates object hallucination evaluation into a series of binary queries about object presence., while AMBER \cite{wang2024amberllmfreemultidimensionalbenchmark} expands evaluation to multiple hallucination dimensions without relying on LLM-based judgments. HallusionBench \cite{guan2024hallusionbenchadvanceddiagnosticsuite} further diagnoses entangled language hallucination and visual illusion, and THRONE \cite{kaul2025throneobjectbasedhallucinationbenchmark} extends evaluation to free-form generation settings. Subsequent efforts such as H-POPE \cite{pham2024hpopehierarchicalpollingbasedprobing} make the evaluation more fine-grained. These benchmarks have substantially improved the measurement of hallucination behaviors in VLMs. In contrast, our work does not propose a new benchmark; instead, we ask why benchmark accuracy can remain unexpectedly stable even when visual evidence is substantially weakened, localized evidence is removed, or queried entities are semantically altered.

\paragraph{Language priors and visual dependence in VLMs.}
A complementary line of research examines whether strong VLM performance truly reflects reliance on image evidence. VLind-Bench \cite{lee-etal-2025-vlind} explicitly studies language priors in large vision--language models, while other recent work questions whether stronger object grounding necessarily reduces hallucination \cite{geigle2024doesobjectgroundingreally} and analyzes the sources of visual object hallucination more directly \cite{jing2025comprehensiveanalysisvisualobject, kogilathota2026halp}. These studies suggest that model outputs may be supported by language-side regularities or other non-grounded signals, even when evaluation scores remain high. Our work is closely related in spirit, but differs in emphasis: rather than proposing a new prior metric or analyzing hallucination only as an output phenomenon, we investigate how benchmark behavior changes under systematic weakening and manipulation of visual evidence, and what such robustness implies about the interpretation of benchmark accuracy.

\paragraph{Fine-grained perception beyond benchmark accuracy.}
Recent studies have also shown that aggregate benchmark performance does not necessarily imply strong visual perception. MME \cite{fu2025mmecomprehensiveevaluationbenchmark} highlights the breadth of multimodal evaluation, while \emph{Eyes Wide Shut?}~\citep{tong2024eyeswideshutexploring} reveals persistent visual shortcomings in multimodal LLMs, and \emph{Do You See Me}~\citep{kanade2025multidimensionalbenchmarkevaluating} further demonstrates that models may perform well on downstream tasks while still exhibiting substantial perception errors. This perspective is closely aligned with our central claim: standard benchmark accuracy can overestimate fine-grained visual grounding, because correct predictions may be sustained by coarse visual cues, representational redundancy, or language priors rather than faithful use of detailed visual evidences \cite{hou2025visionlanguagemodelsreallyunderstand}.

\paragraph{Vision-Language Model Preliminaries}

We consider a vision–language model~\cite{liu2023visualinstructiontuning, sheta2025behavioral} that takes an image $I$ and a text query $Q$ as input and generates a textual response $Y$. The image $I$  is first processed by a vision encoder (\eg, CLIP~\cite{radford2021learningtransferablevisualmodels}) with $L_e$ layers, producing visual token representations $U^{(\ell)} = \{\mathbf{u}_1^{(\ell)}, \dots, \mathbf{u}_M^{(\ell)}\}$ with $\mathbf{u}_i^{(\ell)} \in \mathbb{R}^{d_v}$ at each layer $\ell$. The final encoder output is denote by $U = U^{(L_e)}$. These visual tokens are then projected into the language model input space yielding a sequence of projected visual tokens $V = \{\mathbf{v}_1, \dots, \mathbf{v}_M\}$, where $\mathbf{v}_i \in \mathbb{R}^{d}$. The sequence $V$ is concatenated with the tokenized query to form a multimodal sequence processed by a Transformer decoder~\cite{vaswani2023attentionneed}, and the model autoregressively generates an output sequence.

\begin{figure}[t]
    \centering
    \includegraphics[width=0.47\textwidth]{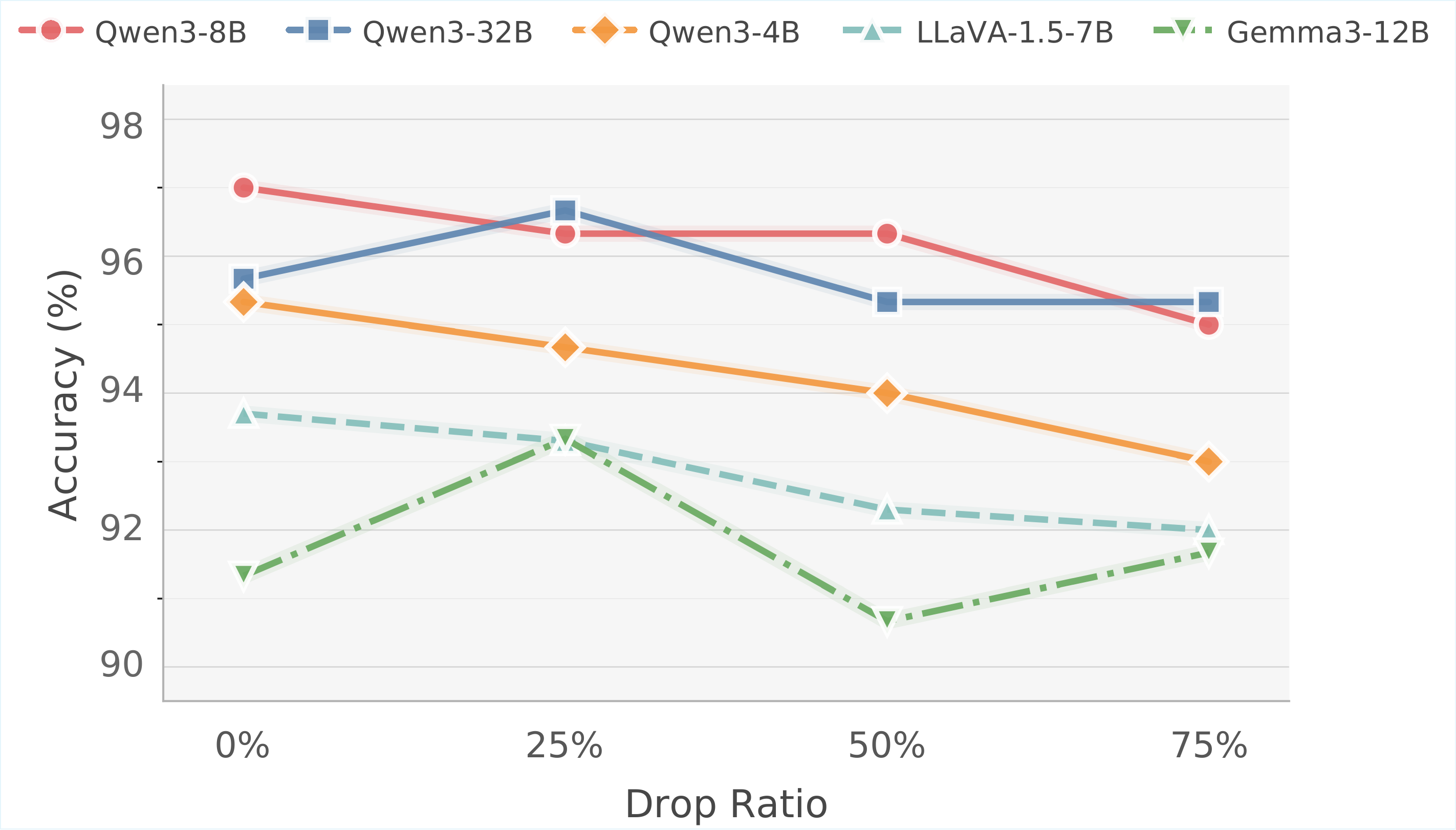}
    \caption{Effect of random image token dropping on POPE accuracy. Even at 75\% token removal, accuracy remains nearly unchanged, suggesting that high performance on this benchmark does not require complete visual representations.}
    \label{fig:token_drop}
\end{figure}

\section{Vision is not Needed}
We conduct standard evaluation of VLM hallucinations on a widely used benchmark POPE~\cite{li2023evaluatingobjecthallucinationlarge}, a benchmark specifically designed to test object hallucination in vision--language models. Its central goal is to examine whether a model claims to see objects that are not actually present in the image. Each sample consists of an image paired with a yes/no question about a queried object or visual concept. A correct response therefore requires the model to verify the queried content against the visual input. As shown in Figure~\ref{fig:token_drop}, we randomly drop image tokens; specifically, we define a drop ratio $\sigma\in[0,1]$, randomly sample a subset of image tokens $S\subset V$, where $(|S|/|V|=\sigma)$, then delete the remaining tokens $V \setminus S$ before passed them to the language decoder $\mathrm{Dec(V) \to \mathrm{Dec(S)}}$. Surprisingly, we find the model performance does not suffer from obvious loss of visual information. For Qwen3-4B and LLaVA-1.5-7B, accuracy decreases approximately linearly as the drop ratio increases, but the magnitude of the drop is small: even when drop ratio $\sigma =0.75$, performance decreases by only about \textbf{3\%} compared to the baseline. In contrast, Qwen3-32B and Gemma3-12B do not exhibit a monotonic decline with increasing token removal. Notably, when $\sigma = 0.25$ both models slightly outperform their baseline accuracy. This raises a more fundamental question: given benchmark scores retains under such severe visual degradation, whether these scores truly reflect the model's reliance on visual evidence.\footnote{One caveat is that the visual tokens used in this intervention are taken after the vision encoder, where bidirectional attention may already have mixed information across patches. As a result, removing a subset of tokens at this stage may not fully eliminate the corresponding visual content. However, our later experiments that intervene directly on the input image, before such encoding and mixing occur, lead to qualitatively similar results. Later, we show that the same pattern remains even when the actual visual evidence is removed from the input image.}

\vspace{-2mm}

\section{Experiments Settings}

To examine whether the above phenomena generalize across tasks and evaluation formats, we study multiple vision--language evaluation settings, including POPE, A-OKVQA, MME, and AMBER~\cite{li2023evaluatingobjecthallucinationlarge,schwenk2022aokvqabenchmarkvisualquestion,fu2025mmecomprehensiveevaluationbenchmark,wang2024amberllmfreemultidimensionalbenchmark}. Among them, POPE serves as the main test point of this work. A-OKVQA and MME are used to test whether the phenomenon shared in different benchmarks, while AMBER provides a complementary view under open-generation evaluation.
On the model side, we evaluate a diverse set of open-source vision--language models, including LLaVA-1.5-7B, Qwen3-VL-{4B, 8B, 32B}, Gemma-3-12B, InternVL3-8B, and Molmo-7B-D-0924~\cite{liu2024improvedbaselinesvisualinstruction,bai2025qwen3vltechnicalreport,gemmateam2025gemma3technicalreport,zhu2025internvl3exploringadvancedtraining,deitke2024molmopixmoopenweights}. Detailed settings can be found in supplementary material~\ref{app:appendix_a}.

For POPE, A-OKVQA, and MME, we primarily report \textbf{accuracy} ($\uparrow$). For entity-level interventions on the positive subset of POPE, we report \textbf{Yes Rate} ($\downarrow$), and the \textbf{decision margin} $\Delta$. Under the \texttt{unknown}-option setting (Defined in Sec.\ref{sec:qa}), we report the \textbf{Unknown Rate}, which measures whether the model awareness of uncertainty. For the open-ended reformulation, we report the target entity's \textbf{rank} ($\downarrow$) and \textbf{MRR} ($\uparrow$, Mean Reciprocal Rank), which measures how highly the correct entity is ranked in the model’s token-level generation distribution. For AMBER, we follow the official evaluation and report \textbf{CHAIR} \cite{rohrbach2019objecthallucinationimagecaptioning}, which measures how often generated responses mention objects not supported by the image, and \textbf{Hallucination} ($\downarrow$), which measures the proportion of responses that contain at least onehallucinated object. We also report \textbf{Coverage}, which measures how much annotated image content is covered by the response, and \textbf{Cog} ($\downarrow$) \cite{wang2024amberllmfreemultidimensionalbenchmark}, which measures how often generated objects fall into a predefined set of human-like hallucinatory targets.

\section{Diagnosing Visual Grounding}

To investigate this question, we conduct interventions at multiple levels. At the visual level, we apply structured degradations to the image input. At the textual level, we perturb the questions and choices. Beyond behavioral observations, we further analyze internal model signals to examine how visual information is processed.

\subsection{Global-Level Visual Interventions}
\label{sec:global}

\paragraph{Setup}
On human visible image intervention, we design a series of ablation experiments at varying granularities as shown in figure~\ref{fig:teaser}. We apply whole-image degradations that systematically weaken the available visual evidence. We consider three settings: \texttt{no-image}, where the image is removed entirely; \texttt{black}, where a proportion $r \in [0,1]$ of the image area is occluded with black masking; and \texttt{blur}, where the original image $I$ is mixed with random noise $\epsilon$,
\begin{equation}
    \tilde{I} = (1 - \alpha)\, I + \alpha\, \epsilon,
\end{equation}
with $\alpha \in [0,1]$ controlling the noise strength. Given this binary answer setting, our evaluate using standard accuracy. If model performance genuinely depends on fine-grained visual evidence, accuracy should degrade proportionally to the severity of corruption. Limited degradation under severe corruption would indicate that surface accuracy is not a sensitive measure of visual dependence.

\paragraph{Results}
\label{sec:result_global}

Table~\ref{tab:global_acc} reports accuracy across seven models on POPE, A-OKVQA, and MME under each global degradation condition. All models drop to near chance-level accuracy under \texttt{no-image}, confirming that they are not independent of visual input. However, under \texttt{black} and \texttt{blur}, accuracy decreases disproportionately little relative to the severity of visual corruption, with most models remaining well above chance level. This pattern holds consistently across all three benchmarks. These results suggest that benchmark accuracy is insensitive to severe visual degradation, indicating a measurement gap between accuracy and true visual dependence.

\subsection{Entity-Level Interventions}
\label{sec:entity}
Global interventions cannot determine whether the model depends on specific visual evidence relevant to a given question. To tackle this, we conduct entity-level manipulations.

\paragraph{Setup}
These interventions require the queried entity to be physically present in the image, and we could then mask or replace such an entity to create a intervention. We therefore restrict an evaluation set with only Ground Truth positive (GT=Yes) samples from POPE, that said the image is guaranteed to contain the entity mentioned in the question. As all samples carry a positive ground truth, we report yes rate — the proportion of responses affirming the entity's presence — as the evaluation metric.

For each question–image pair $(Q,I)$, we extract the target entity from the question using GPT-5 \cite{singh2025openaigpt5card}, localize it with Grounding DINO~\cite{liu2024groundingdinomarryingdino}, and obtain a segmentation mask via SAM2~\cite{ravi2024sam2segmentimages}. Based on these outputs, we construct three interventions of increasing specificity: \texttt{BlackMask} masks only the segmented entity pixels, yielding precise entity-level removal; \texttt{BlackBox} occludes the entire detected bounding box, providing coarser region-level removal that also removes surrounding local context; and \texttt{EntitySwap} replaces the target entity in the image with an image-irrelevant alternative using Gemini-3~\cite{geminiteam2025geminifamilyhighlycapable}, while keeping the question unchanged. Under these adjustments, the \textbf{Yes Rate} becomes the lower the better.

\paragraph{Results}

Table~\ref{tab:comprehensive_entity_occlusion} reports entity-level occlusion results. Across all three benchmarks, both \texttt{BlackMask} and \texttt{BlackBox} lead to \textbf{Yes Rate} drops, confirming that models depend on question-relevant local visual evidence. \texttt{BlackBox} consistently causes a larger drop than \texttt{BlackMask} aligns that \texttt{BlackBox} removes both the entity and its surrounding context (silhouette), whereas \texttt{BlackMask} removes only the entity pixels themselves. This gap indicates that the support for correct predictions is distributed not only over the target entity but also over the local context surrounding it. Notably, even under \texttt{BlackMask}, where the entity has been completely removed, most models still maintain high \textbf{Yes Rate} in POPE, suggesting that surrounding scene context alone may be sufficient to sustain correct predictions.

Table~\ref{tab:entity_swap_accuracy_gemini} reports the results of \texttt{EntitySwap}, which provides a stronger counterfactual test by replacing the queried entity with an unrelated object. Under this condition, the correct answer should always be \textbf{No}, and the ideal \textbf{Yes Rate} should therefore be 0. Yet all models remain well above zero, indicating that they fail to sufficiently update their predictions after the queried entity have changed.

\begin{table*}[t]
\centering
\small
\setlength{\tabcolsep}{10pt}
\renewcommand{\arraystretch}{0.65}
\caption{\textbf{Comprehensive Entity-level Occlusion Results.} We report the Yes Rate(lower is better for Entity-level Interventions conditions ) for each benchmark under three distinct conditions: \textbf{Original} (baseline), \textbf{Mask} (precise entity-level segmentation occlusion), and \textbf{Box} (coarse bounding box occlusion). This comprehensive comparison highlights the varying sensitivity of perception-based and knowledge-based tasks to the loss of local visual features and surrounding context.}
\label{tab:comprehensive_entity_occlusion}
\begin{tabular}{l | ccc | ccc | ccc}
\toprule
\multirow{2}{*}{\textbf{Model}} & \multicolumn{3}{c|}{\textbf{POPE}} & \multicolumn{3}{c|}{\textbf{A-OKVQA}} & \multicolumn{3}{c}{\textbf{MME}} \\
\cmidrule(lr){2-4} \cmidrule(lr){5-7} \cmidrule(lr){8-10}
& Original & Mask & Box & Original & Mask & Box & Original & Mask & Box \\
\midrule

Qwen3-VL-32B    & 0.96 & 0.90 & 0.74 & 0.90 & 0.80 & 0.74 & 0.83 & 0.59 & 0.28 \\

InternVL3-8B    & 0.97 & 0.80 & 0.57 & 0.88 & 0.77 & 0.71 & 0.83 & 0.64 & 0.40 \\

Qwen3-VL-8B     & 0.99 & 0.94 & 0.84 & 0.89 & 0.73 & 0.68 & 0.85 & 0.65 & 0.41 \\

Gemma3-12B   & 0.96 & 0.83 & 0.59 & 0.84 & 0.74 & 0.67 & 0.79 & 0.59 & 0.45 \\

Qwen3-VL-4B     & 0.93 & 0.75 & 0.44 & 0.86 & 0.73 & 0.69 & 0.80 & 0.60 & 0.35 \\

LLaVA-1.5-7B & 0.97 & 0.87 & 0.71 & 0.77 & 0.68 & 0.65 & 0.74 & 0.48 & 0.20 \\

Molmo-7B-D-0924  & 0.94 & 0.67 & 0.46 & 0.82 & 0.66 & 0.60 & 0.82 & 0.54 & 0.27 \\

\bottomrule
\end{tabular}
\end{table*}

\paragraph{Decision Margin Analysis}

\begin{figure}[t]
    \centering
    \includegraphics[width=0.95\linewidth]{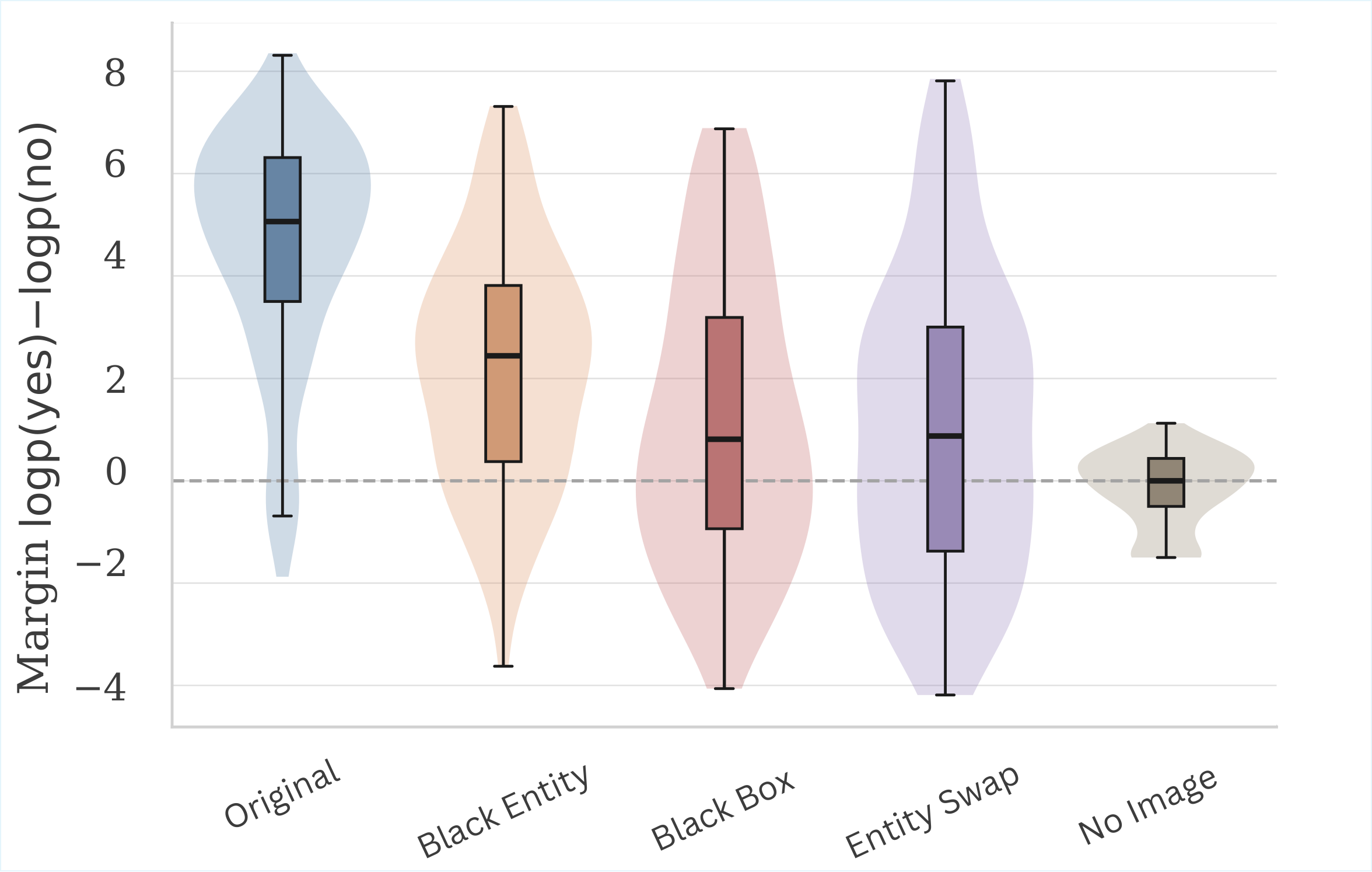}
    \caption{Distribution of decision margins under different visual conditions.Each violin and box summarizes the samples decision margin at the first answer token. The horizontal line at $0$ marks equal preference between the two decisions. As visual evidence is weakened, the margin distribution shifts leftward and becomes less concentrated.
Notably, \textsc{Entity Swap} remains systematically more favorable to ``yes'' than \textsc{No Image}.}
    \label{fig:vilion_margin}
\end{figure}

Move beyond binary predictions, to obtain a finer-grained view, we further examine the decision margin $\Delta = \log p(\texttt{yes}) - \log p(\texttt{no})$,
which captures the model's probability-level preference.
Figure~\ref{fig:vilion_margin} visualizes the distribution of $\Delta$ on the GT=Yes subset under each intervention condition, while detailed numerical statistics are provided in supplementary material~\ref{app:additional_results}.

First, the \texttt{Original} setting shows the largest positive margins, indicating strong affirmative preference when full visual evidence is available and \texttt{No Image} condition indicates little directional preference in the absence of visual input. Second, the margin decreases under all entity-level interventions, confirming that local visual evidence indeed affect the model's decision tendency. However, this decrease is only partial: even under \texttt{Black Entity}, where the queried entity has been removed, the distribution remains clearly shifted toward positive values rather than collapsing toward the \texttt{No Image} condition. This suggests that affirmative preference is not determined by the target entity alone.

A similar pattern holds for \texttt{Black Box} and \texttt{Entity Swap}. Although both conditions reduce the margin further, their distributions remain substantially above \texttt{No Image}.

Overall, the distributional evidence suggests that while models do use entity-relevant visual information, their affirmative bias can still be sustained by broader scene-level visual cues even when the queried entity is removed or semantically replaced. This helps explain why benchmark performance may remain relatively stable despite substantial degradation of fine-grained visual evidence.

\paragraph{Summary}

Taken together, these results suggest that models are not entirely insensitive to local visual evidence, but their dependence on such evidence is not tightly anchored to the queried entity itself. Importantly, although entity-level interventions do alter the model's underlying decision support, these changes are often not reflected in top-1 accuracy. In other words, the model's prediction may remain unchanged even when its internal support for the correct answer has already been substantially weakened.

\begin{table}[t]
\centering
\small
\setlength{\tabcolsep}{8pt}
\renewcommand{\arraystretch}{0.85}
\caption{\textbf{Entity Swap Accuracy Test on Gemini-3 Generated Images.}
We report the Yes rate of different VLMs when the queried entity is replaced with a different entity. Under ideal grounding behavior, the Yes rate should be 0.}
\label{tab:entity_swap_accuracy_gemini}
\begin{tabular}{l c}
\toprule
\textbf{Model} & \textbf{Yes Rate $\downarrow$} \\
\midrule
LLaVA-1.5-7B   & 0.63 \\
Gemma-3-12B    & 0.50 \\
InternVL3-8B       & 0.37 \\
Qwen3-VL-32B      & 0.35 \\
Qwen3-VL-8B       & 0.34 \\
Qwen3-VL-4B       & 0.27 \\
\bottomrule
\end{tabular}
\end{table}

\begin{table}[t]
\centering
\small
\setlength{\tabcolsep}{3pt}
\renewcommand{\arraystretch}{0.75}
\caption{\textbf{Unknown selection rate under different visual conditions.}
We report the fraction of samples for which each model selects the explicit \texttt{unknown} option. Most models remain largely insensitive to the added option, while a few show elevated abstention under stronger visual degradation.}
\label{tab:unknown_rate}
\begin{tabular}{lcccc}
\toprule
\textbf{Model} & \textbf{Normal} & \textbf{No Image} & \textbf{Black (0.75)} & \textbf{Noise (0.75)} \\
\midrule
LLaVA-7B      & 0.00 & 0.00 & 0.00 & 0.00 \\
Gemma-12B     & 0.03 & 0.00 & 0.09 & 0.78 \\
Qwen3-32B   & 0.01 & 0.00 & 0.05 & 0.81 \\
Qwen3-8B   & 0.00 & 0.00 & 0.00 & 0.00 \\
Qwen3-4B   & 0.00 & 0.00 & 0.00 & 0.03 \\
InternVL3-8B   & 0.00 & 0.11 & 0.00 & 0.01 \\
Molmo-7B      & 0.01 & 0.00 & 0.02 & 0.16 \\
\bottomrule
\end{tabular}
\end{table}

\begin{table*}[t]
\centering
\small
\setlength{\tabcolsep}{10pt}
\renewcommand{\arraystretch}{0.65}
\caption{\textbf{Ranking and Probability Metrics under Entity-level Interventions.} We report the Mean Reciprocal Rank (MRR), Mean/Median Rank, and Mean Prediction Probability across 117 samples. This comparison illustrates the severe degradation in the model's retrieval and ranking capabilities when specific visual entities are perturbed.}
\label{tab:ranking_metrics_comparison}
\begin{tabular}{l | cc | cc | c}
\toprule
\multirow{2}{*}{\textbf{Condition}} & \multicolumn{2}{c|}{\textbf{Mean Reciprocal Rank (MRR)} $\uparrow$} & \multicolumn{2}{c|}{\textbf{Rank Analysis} $\downarrow$} & \multirow{2}{*}{\textbf{Mean Probability} $\uparrow$} \\
\cmidrule(lr){2-3} \cmidrule(lr){4-5}
& All Samples & Valid Only & Mean Rank & Median Rank & \\
\midrule
Original (Baseline) & 0.0322 & 0.0322 & 391.91 & 85 & 0.000690 \\
Perturbed (Entity Swap) & 0.0101 & 0.0101 & 942.98 & 458 & 0.000142 \\
\midrule
\rowcolor[gray]{0.95} \textbf{Relative Change} & \textbf{-68.6\%} & \textbf{-68.6\%} & \textbf{+140.6\%} & \textbf{+438.8\%} & \textbf{-79.4\%} \\
\bottomrule
\end{tabular}
\end{table*}

\begin{table*}[t]
\centering
\small
\caption{Overall Accuracy (Acc) across three benchmarks under various visual perturbations. \textbf{Normal} represents the vanilla setting, while \textbf{No Image}, \textbf{Black}, and \textbf{Blur} denote different levels of visual degradation.}
\label{tab:global_acc}
\setlength{\tabcolsep}{12pt}
\renewcommand{\arraystretch}{0.70}
\begin{tabular*}{\textwidth}{@{\extracolsep{\fill}}llcccccc}
\toprule
\multirow{2}{*}{\textbf{Model}} & \multirow{2}{*}{\textbf{Dataset}} & \textbf{Normal} & \textbf{No Image} & \multicolumn{2}{c}{\textbf{Black}} & \multicolumn{2}{c}{\textbf{Blur}} \\
\cmidrule(lr){5-6} \cmidrule(lr){7-8}
& & (Baseline) & (Zero) & $p=0.5$ & $p=0.75$ & $p=0.5$ & $p=0.75$ \\
\midrule

\multirow{3}{*}{Qwen3-VL-32B}
& Pope & 0.96 & 0.50 & 0.87 & 0.81 & 0.89 & 0.60 \\
& A-OKVQA & 0.90 & 0.51 & 0.79 & 0.72 & 0.78 & 0.73 \\
& MME & 0.92 & 0.54 & 0.86 & 0.81 & 0.85 & 0.81 \\
\midrule

\multirow{3}{*}{InternVL3-8B}
& Pope & 0.98 & 0.50 & 0.86 & 0.80 & 0.93 & 0.71 \\
& A-OKVQA & 0.88 & 0.48 & 0.79 & 0.71 & 0.77 & 0.70 \\
& MME & 0.89 & 0.56 & 0.82 & 0.78 & 0.82 & 0.80 \\
\midrule

\multirow{3}{*}{Qwen3-VL-8B}
& Pope & 0.97 & 0.50 & 0.86 & 0.78 & 0.91 & 0.71 \\
& A-OKVQA & 0.89 & 0.47 & 0.77 & 0.69 & 0.76 & 0.70 \\
& MME & 0.88 & 0.52 & 0.81 & 0.76 & 0.84 & 0.79 \\
\midrule

\multirow{3}{*}{Gemma-3-12B}
& Pope & 0.94 & 0.57 & 0.89 & 0.80 & 0.78 & 0.54 \\
& A-OKVQA & 0.84 & 0.46 & 0.74 & 0.68 & 0.70 & 0.66 \\
& MME & 0.83 & 0.56 & 0.78 & 0.74 & 0.75 & 0.73 \\
\midrule

\multirow{3}{*}{Qwen3-VL-4B}
& Pope & 0.95 & 0.50 & 0.87 & 0.77 & 0.90 & 0.65 \\
& A-OKVQA & 0.86 & 0.47 & 0.76 & 0.68 & 0.75 & 0.70 \\
& MME & 0.81 & 0.51 & 0.75 & 0.70 & 0.76 & 0.73 \\
\midrule

\multirow{3}{*}{LLaVA-1.5-7B}
& Pope & 0.94 & 0.50 & 0.87 & 0.76 & 0.90 & 0.72 \\
& A-OKVQA & 0.77 & 0.38 & 0.69 & 0.65 & 0.69 & 0.64 \\
& MME & 0.80 & 0.50 & 0.73 & 0.69 & 0.74 & 0.72 \\
\midrule

\multirow{3}{*}{Molmo-7B-D-0924}
& Pope & 0.95 & 0.51 & 0.85 & 0.77 & 0.92 & 0.75 \\
& A-OKVQA & 0.82 & 0.47 & 0.69 & 0.64 & 0.70 & 0.64 \\
& MME & 0.79 & 0.51 & 0.71 & 0.66 & 0.74 & 0.71 \\

\bottomrule
\end{tabular*}
\end{table*}

\subsection{Task-Formulation Interventions}
\label{sec:qa}

The preceding interventions perturb the visual input itself, but retain the original closed-form question formats (yes/no and multiple-choice) which may allow models to exploit answer-format priors without genuinely engaging with the visual content. To reduce this confound, we modify the question formulation in two ways to move beyond closed-form answers.

\paragraph{Interventions on Original Question Formulation}
\textbf{First}, under the global degradation settings in Section~\ref{sec:global}, we extend the original yes/no task with an explicit \texttt{unknown} option. The purpose is to provide a backup buffer for VLMs when visual evidence is insufficient to support a confident binary judgment. We report the \textbf{Unknown Rate} to measure whether the model can express uncertainty under degraded visual conditions. \textbf{Second}, we reformulate the original object-presence questions as open-ended generation tasks, asking the model to answer which objects are present in the image. We evaluate the rank of the target entity in the output distribution and report its \textbf{MRR}(Mean Reciprocal Rank, Tab. ~\ref{tab:ranking_metrics_comparison}), which measures how highly the correct entity is ranked in the model's token-level generation distribution. If the queried entity is captured, it should appear with higher salience in the generation distribution.

\paragraph{Results on Original Question Formulation}

After introducing an explicit \texttt{unknown} option (Table \ref{tab:unknown_rate}), most models still rarely choose \texttt{unknown} under degraded conditions. Even when visual evidence is extremely weak or entirely absent, they generally continue to produce definite binary judgments. This suggests that the observed output stability is not simply imposed by the original yes/no answer space. Even when given a softer fallback, current models still show limited ability to express uncertainty in proportion to insufficient evidence. When the original object-presence questions are reformulated as open-ended generation, the target entity receives a low rank in the output distribution (Tab.~\ref{tab:ranking_metrics_comparison}). The model's representation of the target entity is intrinsically weak: even when the entity is visible, it is not strongly encoded in the generation distribution. This provides a representational explanation for why entity-level interventions have limited impact on accuracy---the model never strongly relies on entity-specific information in making its decision.

\paragraph{Complementary Open-Generation}

The reformulations above still operate on questions derived from the original benchmark. To complement these derived open-generation tasks, we additionally evaluate on AMBER \cite{wang2024amberllmfreemultidimensionalbenchmark}, a benchmark natively designed for open-ended visual description. We follow the official AMBER evaluation protocol and introduce only coarse-grained visual interventions to the input images. Compared with closed-form tasks, this setting more directly evaluates whether the model can generate object-level content consistent with the image.

\paragraph{Results on the Complementary Open-Generation}

We observe the same trend under the native open-generation setting of AMBER (Tab.~\ref{tab:amber_gen_results}). Under degradation, hallucination increases and coverage decreases, but the overall changes remain limited; only \texttt{no-image} causes drastic deterioration. This shows that the relative stability observed above is not merely an artifact of the yes/no format, but a consistent phenomenon across evaluation settings.

\section{Representational Analysis}
\subsection{Layer-wise Analysis of Vision Token Geometry}
\label{sec:geometry}

The above experiments examine model behavior by manipulating inputs and outputs. Here we complement behavioral evidence with a representational analysis \cite{sheta2025behavioral, li2026on}, asking whether the vision encoder itself progressively loses the spatial discriminability required for fine-grained grounding. For each encoder layer $\ell \in \{1, \dots, L_e\}$, we impose a regular $4 \times 4$ spatial partition over the visual tokens $U^{(\ell)}$ and compute three complementary metrics.

First, we measure \textit{intra-block} and \textit{inter-block} \textbf{cosine similarity}—the mean pairwise cosine between tokens within the same spatial block and across different blocks. If inter-block similarity increases with depth, tokens from different spatial regions become directionally indistinguishable. Second, we apply $k$-means clustering ($k{=}16$, matching the $4 \times 4$ partition) to token representations at each layer and evaluate how well the clusters align with the ground-truth spatial blocks (\textit{k-means spatial compactness}). Third, we compute the \textit{effective rank}~\cite{7098875} of the token representation matrix:
\begin{equation}
    \text{erank}(U^{(\ell)}) = \exp\!\Bigl(-\sum_j \bar{\sigma}_j \log \bar{\sigma}_j\Bigr),
\end{equation}
where $\bar{\sigma}_j = \sigma_j / \sum_k \sigma_k$ are the normalized singular values of $U^{(\ell)}$. A decrease in effective rank indicates that token representations collapse onto a lower-dimensional subspace.

Together, these metrics characterize spatial discriminability from complementary perspectives: directional alignment, spatial separability, and representational dimensionality. If all three degrade in deeper layers, this would provide a representational explanation consistent with the behavioral finding that model predictions are insufficiently sensitive to the loss of fine-grained local evidence.

\subsection{Spatial Discriminability Degrades with Depth}
\label{sec:result_geometry}

\begin{figure*}[htbp]
    \centering
    \includegraphics[width=0.95\textwidth]{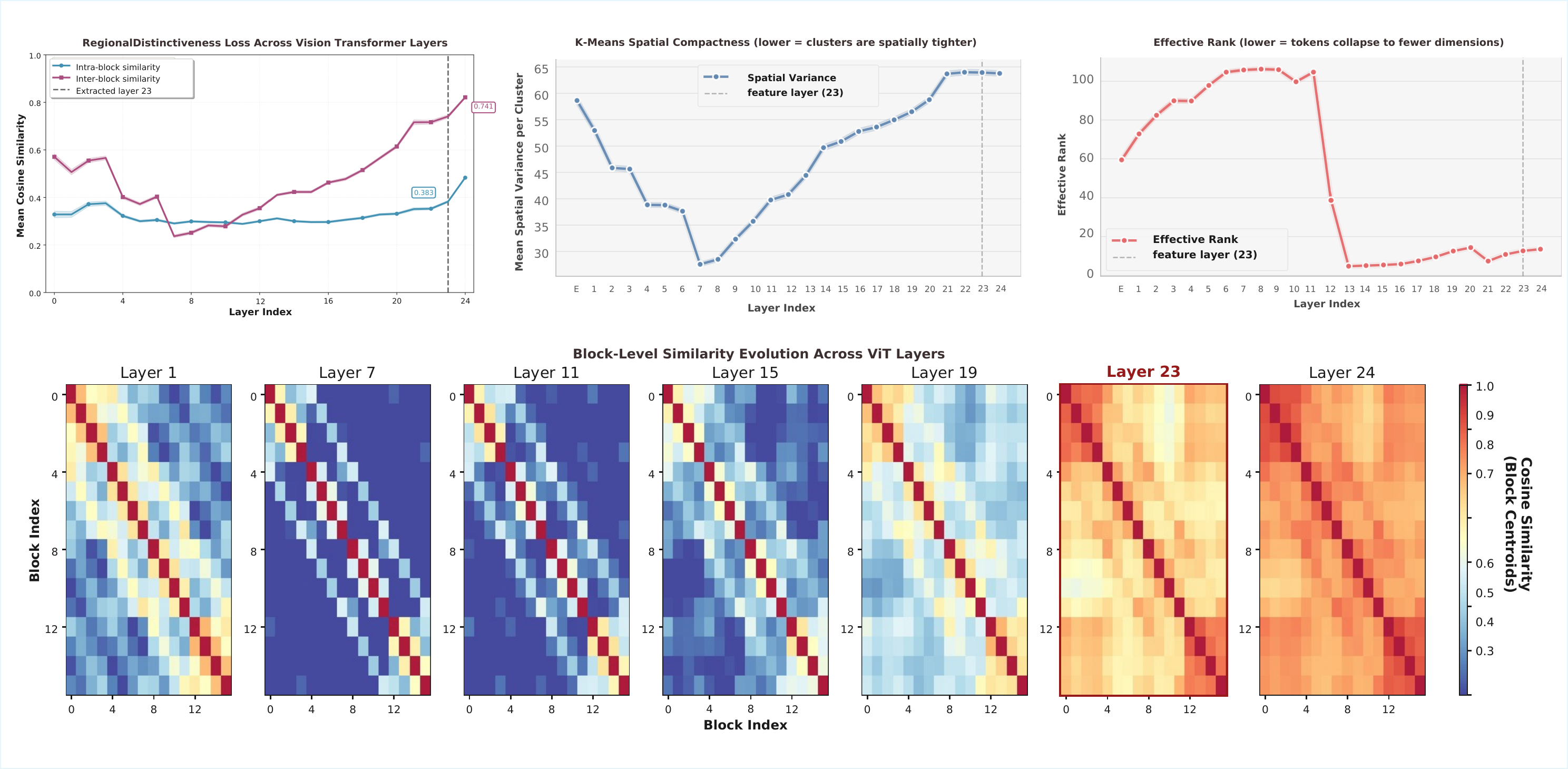}
    \caption{Layer-wise representational analysis of visual tokens in the vision encoder. We evaluate spatial discriminability from three complementary perspectives: intra-/inter-block cosine similarity, $k$-means spatial compactness, and effective rank. These results provide a possible representational explanation for the limited sensitivity of model predictions to the loss of fine-grained local evidence.}
    \label{fig:VIT_analysis}
\end{figure*}

The \textbf{Top left panel} of Figure~\ref{fig:VIT_analysis} reports the layer-wise evolution of block-wise cosine similarity. In early layers, intra-block similarity exceeds inter-block similarity, indicating that tokens within the same spatial region remain more similar than those across regions. As depth increases, inter-block similarity rises steadily and approaches intra-block similarity, substantially shrinking the gap between them. The \textbf{heatmaps} in Figure~\ref{fig:VIT_analysis} visualize this trend directly: early layers show low off-diagonal values, reflecting strong differences between spatial blocks, whereas by Layer 23 the matrix becomes nearly uniform, indicating that spatial distinctions have largely disappeared.

Notably, around Layers 7–9, inter-block similarity briefly falls below intra-block similarity, suggesting that local spatial structure is strengthened in intermediate layers before the later convergence observed in deeper layers.

Figure~\ref{fig:VIT_analysis} also reports \textbf{k-means spatial compactness} and \textbf{effective rank} across layers. Spatial variance reaches its minimum near Layer 7, indicating that clustering at this depth best preserves spatial structure; beyond this point, variance increases steadily, showing that deeper-layer clusters no longer align with spatial regions. Effective rank remains high through the first 12 layers and then drops sharply, indicating that token representations collapse onto a much lower-dimensional subspace.

\paragraph{Summary} Across all three metrics 1)directional alignment, 2)spatial separability, and 3)dimensional diversity,the same pattern emerges: spatial discriminability progressively degrades in deeper layers. This representation-level trend is consistent with the behavioral findings: if tokens from different spatial regions become mixed in the encoder, downstream computation is less likely to maintain strong dependence on fine-grained local visual evidence.

\section{Discussion}
\label{sec:discussion}

\subsection{Visual Reliance in VLMs Is Shallow}

Our results indicate that the visual reliance of current vision--language models is shallow and rarely grounded in fine-grained entity-level evidence. Across global degradation experiments, substantial corruption of visual input leads to only minor drops in benchmark accuracy, suggesting that coarse scene-level cues are often sufficient to sustain correct predictions. This pattern becomes more evident under entity-level interventions. Even after the queried entity is removed, models frequently remain confidently affirmative, indicating that surrounding scene context can substitute for direct visual evidence of the entity itself. Our probability-level analysis further supports this observation.

Additional analysis reinforces this conclusion. MRR results show that even when the queried entity is present, models often fail to assign it high priority in the generation distribution. Our representational analysis provides a possible explanation for this shallow reliance.
As visual tokens propagate through deeper layers of the encoder, spatial discriminability between regions progressively degrades. Such representational homogenization may reduce the availability of precise local evidence required for entity-level grounding.

Taken together, these findings suggest that correct predictions in current VLM benchmarks often do not require precise visual evidence.
\subsection{Current Benchmarks Fail to Capture This Problem}

If models can produce correct answers without perceiving the queried entity, benchmark accuracy alone cannot faithfully reflect visual grounding capability.  This limitation can be vital if accuracy is used for VLM research with a take-as-granted assumption. If benchmarks cannot reliably measure fine-grained grounding, methods intended to improve visual perception, mitigate hallucination, or reduce image tokens for inference efficiency may appear effective simply because they exploit coarse scene cues or dataset priors. As a result, both the effectiveness of such methods and the visual capability of existing VLMs may be misestimated. Our results suggest that evaluation should move beyond top-1 accuracy and explicitly test whether model predictions respond appropriately to changes in visual evidence. Evaluation should test whether predictions change appropriately under visual perturbations.

\section{Conclusion and Future Work}
\label{sec:conclusion}

Across multi-granularity interventions, we find that benchmark performance remains stable even when visual evidence is weakened. At the same time, probability-level analyses reveal that model predictions become less confident, and entity-level evaluations show that outputs are weakly anchored to the queried entity itself. These results suggest that standard benchmarks do not faithfully reflect fine-grained visual grounding. In particular, models can maintain correct predictions under misleading visual inputs, indicating that current benchmarks may overestimate visual reliance. Many research directions---including hallucination mitigation, visual token pruning, and claims about improved grounding---are evaluated primarily through theses benchmark . If such metrics are insensitive to the loss of fine-grained visual evidence, they may provide misleading signals about model behavior. In future work, we plan to construct a benchmark that relies purely on visual evidence, aiming to reflect the real visual capabilities of models. On the modeling side, we aim to improve the model's ability to utilize visual evidence, such that it can make correct predictions even when visual inputs are weakened.
\newpage

{
    \small
    \bibliographystyle{ieeenat_fullname}
    \bibliography{main}
}

    \clearpage
\setcounter{page}{1}
\maketitlesupplementary

\setcounter{section}{0}
\renewcommand{\thesection}{\Alph{section}}
\renewcommand{\thesubsection}{\Alph{section}.\arabic{subsection}}
\renewcommand{\thetable}{\Alph{section}.\arabic{table}}
\renewcommand{\thefigure}{\Alph{section}.\arabic{figure}}
\setcounter{table}{0}
\setcounter{figure}{0}

\section{Implementation and Experimental Details}
\label{app:appendix_a}
\subsection{Benchmarks and Data Splits}
\label{app:benchmarks}

We conduct our analysis on four widely used vision-language benchmarks: POPE, A-OKVQA, MME, and AMBER. In the main paper, POPE serves as the primary benchmark for controlled analysis, while A-OKVQA and MME are used to examine whether the observed phenomena extend beyond a single benchmark setting. AMBER is further included as a complementary benchmark for open-generation evaluation.

For POPE, we use a 300-example evaluation set from the official benchmark setting adopted in our study. For A-OKVQA, we evaluate on the validation split and use approximately 1.15K examples. For MME, we follow the full benchmark setting and evaluate on all examples, totaling approximately 2.37K samples. For AMBER, we follow the official evaluation protocol.

Although our study covers multiple benchmarks, the most fine-grained interventions are conducted primarily on POPE. This design is mainly motivated by the substantially higher cost of entity-level semantic manipulation, especially when image editing, quality screening, and manual verification are required. As a result, POPE serves as the main testbed for our most controlled intervention analyses.

\subsection{Model Suite and Inference Setup}
\label{app:models}

We evaluate a diverse set of open-source vision-language models, including LLaVA-1.5-7B, Qwen3-VL-4B, Qwen3-VL-8B, Qwen3-VL-32B, Gemma-3-12B, InternVL3-8B, and Molmo-7B-D-0924, as listed in the main paper.

Unless otherwise noted, all experiments are conducted under a unified greedy decoding setup. We use a consistent prompt format within each benchmark and evaluation condition so that the observed differences are less likely to be caused by prompt variation. Experiments are run on NVIDIA RTX 4090 GPUs using bfloat16 precision.

Our implementation is primarily based on the Hugging Face \texttt{transformers} library. For Molmo, we use \texttt{transformers==4.44.2} due to compatibility issues with newer versions in our inference pipeline.

\subsection{Selection of the POPE Positive Subset}
\label{app:pope_positive_subset}

Several analyses in this work focus specifically on the positive subset of POPE, namely, examples whose ground-truth answer is \textit{Yes}. Starting from the 300-example POPE evaluation set, we identify 149 examples with ground-truth label \textit{Yes}.

This subset is used in analyses where affirmative support is the quantity of interest, such as examining how strongly the model continues to support the correct positive answer under different visual interventions. Restricting the analysis to ground-truth positive examples allows us to track changes in positive evidence more directly, without conflating them with shifts caused by class balance or answer priors.

For the more demanding semantic-manipulation experiments, the usable subset becomes smaller after additional filtering. In particular, we retain only those examples for which the queried entity can be reliably localized, segmented, semantically modified, and manually verified. This results in a final entity-swap subset of 117 examples.

\subsection{Motivation Example Setup}
\label{app:Motivation}

\paragraph{Models.}
We conduct the experiment on multiple vision--language models, including Qwen3-VL-4B/8B/32B, LLaVA-1.5-7B, and Gemma-3-12B.

\paragraph{Image token dropping.}
For all models, we remove a subset of image tokens at the \emph{embedding level} before they are fed into the LLM.

Concretely, given the input sequence composed of text embeddings and image embeddings,
\[
[\mathbf{E}_{\text{text}}, \mathbf{E}_{\text{img}}],
\]
we randomly drop a portion of image embeddings $\mathbf{E}_{\text{img}}$ with drop ratio $r \in \{0, 0.25, 0.5, 0.75\}$, and construct a new input sequence:
\[
[\mathbf{E}_{\text{text}}, \mathbf{E}_{\text{img}}^{\text{kept}}].
\]

\paragraph{Where the embeddings come from.}
Image embeddings are taken \emph{after} the visual encoder and projection/merger module, i.e., the final image-token representations that are normally concatenated with text tokens before entering the LLM.

For Qwen3-VL, we explicitly extract these embeddings from the visual merger module. For other models (e.g., LLaVA-1.5, Gemma-3), we apply the same operation at the corresponding image-token embedding stage.

\paragraph{Implementation.}
We identify the positions of image tokens in the input sequence, remove a random subset of them, and directly feed the modified embeddings into the LLM using the \texttt{inputs\_embeds} interface (or an equivalent mechanism).

\paragraph{Inference.}
We use greedy decoding with a maximum of 6 generated tokens. Each sample is processed independently due to variable numbers of image tokens after dropping.

\subsection{Global Visual Degradation Settings}
\label{app:global_degradation}

We evaluate several global visual degradation settings applied directly to the input images before feeding them into the model.

\paragraph{No-image.}
We replace the input image with a dummy black image of fixed size. The model still receives an image input, but it contains no valid visual information. Since some models do not accept inputs lacking an image, we treat the absence of visual information as "no images" to ensure alignment during testing.

\paragraph{Black occlusion.}
We apply structured black occlusion to the image. Specifically, given an image of size $H \times W$:
\begin{itemize}
    \item $p=0.5$: the top half of the image is set to black;
    \item $p=0.75$: the top half and the bottom-left quarter are set to black;
    \item $p=1.0$: the entire image is set to black.
\end{itemize}

\paragraph{Noise corruption.}
We mix the original image with random noise. Given an image $\mathbf{I}$ and random noise $\mathbf{N}$, the corrupted image is:
\[
\mathbf{I}' = (1 - p)\mathbf{I} + p\mathbf{N},
\]
where $p \in \{0.5, 0.75, 1.0\}$ controls the corruption strength.

\paragraph{Implementation.}
All perturbations are applied in pixel space before the standard image preprocessing pipeline. The same transformations are used across all models.

\subsection{Construction of the Entity-Level Evaluation}
\label{app:entity_subset}

We construct an entity-level evaluation subset to support localized visual interventions, including \textit{BlackMask}, \textit{BlackBox}, and \textit{Entity Swap}.

\paragraph{Target extraction and localization.}
For each sample, we first extract the queried entity from the question using a language model. We then localize the entity in the image using Grounding DINO, followed by SAM2 to obtain a pixel-level segmentation mask.

Only samples with a single, clearly identifiable target entity are retained.

\paragraph{Entity-level interventions.}

Given the localized entity, we construct three types of interventions:

\begin{itemize}
    \item \textbf{BlackMask.}
    We mask only the pixels corresponding to the segmented entity region. Specifically, pixels inside the SAM2 mask are set to black, while all other pixels remain unchanged. This removes direct visual evidence of the entity while preserving surrounding context.

    \item \textbf{BlackBox.}
    We occlude the entire bounding box of the entity. Concretely, all pixels inside the Grounding DINO bounding box are set to black. Compared to BlackMask, this removes both the entity and its immediate local context.

    \item \textbf{Entity Swap.}
    We replace the target entity with a different object using an image editing model (Gemini). The replacement is conditioned on the original image and the extracted entity, while keeping the rest of the scene unchanged. The goal is to introduce semantically incorrect but visually plausible content.
\end{itemize}

\paragraph{Filtering.}
Not all samples can be reliably processed through the full pipeline. We apply the following filtering steps:

\begin{itemize}
    \item Remove samples with multiple queried entities;
    \item Remove cases where Grounding DINO fails to localize the target;
    \item Remove cases where SAM2 produces invalid or empty masks;
    \item For Entity Swap, remove cases where the editing model fails to generate a clear and consistent replacement.
\end{itemize}

\paragraph{Final subset.}
Due to failures in detection, segmentation, and image editing, the usable subset is reduced. After automatic filtering, repeated generation, and manual verification, the final entity-swap subset contains 117 samples. The same subset is used across all entity-level interventions.

\subsection{Decision Margin Analysis}
\label{app:decision_margin}

We compute decision margins based on the model's token-level probabilities at the first answer token.

\paragraph{Definition.}
Given the logits at the first generated token, we compute:
\[
\Delta = \log P(\text{yes}) - \log P(\text{no}),
\]
where $P(\text{yes})$ and $P(\text{no})$ are obtained by applying a softmax over the vocabulary.

\paragraph{Implementation.}
For each input, we construct a standard chat-style prompt and perform a forward pass without decoding. We extract the logits corresponding to the next token (i.e., the first answer token position), and compute log-probabilities via \texttt{log\_softmax}.

The token IDs for \texttt{yes} and \texttt{no} are obtained from the tokenizer (including leading whitespace when required), and the margin is computed as the difference between their log-probabilities.

\paragraph{Evaluation protocol.}
Margins are computed independently for each input under different visual conditions (e.g., original, black entity, black box, entity swap, no-image). All samples are processed without sampling, and no generation is performed beyond the first token.

\paragraph{Subset.}
For analysis, we report margin distributions on the GT=Yes subset, following the same data filtering as in the entity-level evaluation.

\paragraph{Visualization.}
We plot the distribution of margins across samples using violin and box plots, with $\Delta = 0$ as the decision boundary.

\subsection{Task-Formulation Interventions}
\label{app:task_intervention}

We introduce two task-level interventions by modifying the answer space and generation format.

\paragraph{Multiple-choice with unknown option.}
In the global visual degradation setting, we extend the binary answer space (yes/no) by adding an explicit \texttt{unknown} option. The prompt is kept unchanged except for the instruction to select from \texttt{yes}, \texttt{no}, or \texttt{unknown}.

\paragraph{Open generation.}
For entity-level analysis, we reformulate the task as open-ended generation. Instead of binary questions, we prompt the model to produce likely objects in the image (e.g., ``List the most likely objects in the image'').

\paragraph{Logit-based entity ranking.}
We evaluate whether the target entity is captured in the model's output distribution using a logit-based ranking.

For each input, we extract the logits at the first generated token and compute the probability distribution over the vocabulary. Let $p(v)$ denote the probability of token $v$. Given a target entity, we obtain its first token ID via the tokenizer, and compute its rank among all tokens sorted by $p(v)$.

The reciprocal rank is defined as:
\[
\mathrm{RR} = \frac{1}{\mathrm{rank}},
\]
and the final metric is the mean reciprocal rank (MRR) over all samples.

\paragraph{Implementation.}
We use greedy decoding with \texttt{max\_new\_tokens=1} and extract the first-step logits via \texttt{output\_scores=True}. Ranking is performed directly on the softmax-normalized logits without generating full text sequences.

\subsection{Spatial Structure and Representation Analysis Setup}
\label{app:spatial_analysis}

We analyze spatial structure and representation properties of visual tokens extracted from the vision encoder.

\paragraph{Visual Tokens.}
For all models, we operate on patch-level visual tokens produced by the vision backbone. For ViT-based encoders, this corresponds to a fixed spatial grid (e.g., $24\times24=576$ tokens per image). The \texttt{[CLS]} token is excluded. Hidden states from all layers are collected for analysis.

\paragraph{Block Partition.}
For block-based analysis, the spatial grid is evenly divided into $4\times4$ non-overlapping regions. Each block contains an equal number of tokens (e.g., $6\times6$ tokens per block for a $24\times24$ grid). Token indices are mapped to 2D coordinates based on row-major ordering.

\paragraph{Similarity Computation.}
All similarity measurements are based on cosine similarity. Token features are $\ell_2$-normalized before computing pairwise similarities. Intra-block similarity is computed over all token pairs within each block. Inter-block similarity is computed between block-level centroids, obtained by averaging token features within each block.

\paragraph{K-Means Clustering.}
For data-driven spatial analysis, we perform K-means clustering on token features at each layer. Tokens are clustered into $K=16$ groups using Euclidean distance. Clustering is performed independently for each image and each layer, with a fixed number of iterations.

\paragraph{Effective Rank.}
To measure representational diversity, we compute the effective rank of the token feature matrix at each layer. Token features are first mean-centered. Singular values are obtained via SVD, and effective rank is computed as the exponential of the entropy of the normalized squared singular values.

\paragraph{Aggregation.}
All metrics are computed per image and per layer, and then averaged across the dataset. Confidence intervals are estimated using standard error across samples.

\section{More Additional Results}
\label{app:additional_results}

\begin{table*}[t]
\centering
\footnotesize
\setlength{\tabcolsep}{4.5pt}
\renewcommand{\arraystretch}{1.10}
\caption{\textbf{AMBER (Generative) results.} Lower is better for CHAIR/Hal/Cog; higher is better for Cover.}
\label{tab:amber_gen_results}
\begin{tabular}{llcccc}
\toprule
Model & Condition & CHAIR $\downarrow$ & Cover $\uparrow$ & Hal $\downarrow$ & Cog $\downarrow$ \\
\midrule

\multirow{6}{*}{Qwen3-8B}
& Original    & 6.8  & 47.6 & 32.8 & 1.3 \\
& Black ($p{=}0.5$)  & 11.5 & 42.4 & 45.7 & 1.1 \\
& Black ($p{=}0.75$) & 9.8  & 36.9 & 33.8 & 1.3 \\
& Blur ($p{=}0.5$)   & 6.8  & 45.3 & 28.2 & 1.1 \\
& Blur ($p{=}0.75$)  & 7.6  & 42.6 & 26.7 & 1.2 \\
& No Image   & 76.6 & 13.4 & 95.5 & 14.8 \\
\midrule

\multirow{6}{*}{Qwen3-4B}
& Original    & 7.2  & 52.3 & 35.6 & 0.9 \\
& Black ($p{=}0.5$)  & 11.7 & 45.6 & 54.4 & 1.3 \\
& Black ($p{=}0.75$) & 9.7  & 39.3 & 38.1 & 1.3 \\
& Blur ($p{=}0.5$)   & 6.3  & 56.1 & 32.0 & 1.5 \\
& Blur ($p{=}0.75$)  & 8.4  & 52.8 & 34.5 & 1.9 \\
& No Image   & 68.9 & 7.9  & 78.8 & 10.1 \\
\midrule

\multirow{6}{*}{Qwen3-32B}
& Original    & 8.1  & 37.6 & 26.9 & 0.8 \\
& Black ($p{=}0.5$)  & 10.8 & 33.5 & 36.7 & 0.9 \\
& Black ($p{=}0.75$) & 12.1 & 26.6 & 30.1 & 1.1 \\
& Blur ($p{=}0.5$)   & 7.9  & 36.3 & 25.5 & 0.9 \\
& Blur ($p{=}0.75$)  & 9.1  & 33.2 & 26.6 & 1.1 \\
& No Image   & 71.1 & 0.4  & 5.1  & 0.8 \\
\midrule

\multirow{6}{*}{LLaVA-1.5-7B}
& Original    & 7.4  & 49.4 & 31.7 & 3.7 \\
& Black ($p{=}0.5$)  & 8.9  & 42.7 & 31.7 & 3.4 \\
& Black ($p{=}0.75$) & 11.3 & 36.5 & 33.5 & 3.2 \\
& Blur ($p{=}0.5$)   & 8.5  & 46.1 & 30.5 & 2.9 \\
& Blur ($p{=}0.75$)  & 10.2 & 44.0 & 31.4 & 3.1 \\
& No Image   & 48.3 & 6.4  & 64.3 & 11.3 \\
\midrule

\multirow{6}{*}{Gemma3-12B}
& Original    & 5.5  & 46.6 & 26.7 & 0.8 \\
& Black ($p{=}0.5$)  & 8.7  & 38.7 & 34.2 & 0.7 \\
& Black ($p{=}0.75$) & 8.5  & 29.6 & 22.1 & 0.5 \\
& Blur ($p{=}0.5$)   & 7.1  & 45.1 & 28.2 & 1.1 \\
& Blur ($p{=}0.75$)  & 10.4 & 41.0 & 33.6 & 1.6 \\
& No Image   & 66.5 & 0.0  & 99.9 & 0.0 \\
\midrule

\multirow{6}{*}{InternVL3-8B}
& Original    & 5.1  & 52.8 & 31.7 & 1.4 \\
& Black ($p{=}0.5$)  & 6.3  & 45.7 & 31.5 & 1.8 \\
& Black ($p{=}0.75$) & 9.1  & 40.5 & 36.1 & 1.8 \\
& Blur ($p{=}0.5$)   & 6.3  & 51.7 & 32.0 & 2.1 \\
& Blur ($p{=}0.75$)  & 7.1  & 46.8 & 33.7 & 2.0 \\
& No Image   & 0.0   & 0.0   & 0.0   & 0.0 \\
\midrule

\multirow{6}{*}{Molmo-7B}
& Original    & 8.4  & 65.4 & 47.0 & 2.6 \\
& Black ($p{=}0.5$)  & 10.6 & 53.7 & 51.3 & 2.5 \\
& Black ($p{=}0.75$) & 11.5 & 43.0 & 44.6 & 2.2 \\
& Blur ($p{=}0.5$)   & 8.2  & 63.6 & 41.4 & 2.6 \\
& Blur ($p{=}0.75$)  & 10.6 & 59.9 & 48.2 & 3.4 \\
& No Image   & 0.0  & 0.0  & 0.0  & 0.0 \\

\bottomrule
\end{tabular}
\end{table*}

\subsection{Detailed Results on AMBER}
\label{app:amber_results}

We first report detailed results on AMBER, an open-generation benchmark that complements the closed-form evaluations in the main paper. Because the full AMBER results table is relatively large and involves multiple generative evaluation metrics, we place it in the appendix for completeness.

\begin{table*}[t]
\centering
\small
\setlength{\tabcolsep}{4.7pt}
\caption{
Detailed POPE results with expanded answer options.}
\label{tab:unknown_detailed_results}
\begin{tabular}{llccccccccc}
\hline
\textbf{Model}
& \textbf{Condition}
& \textbf{Acc} $\uparrow$
& \textbf{Prec} $\uparrow$
& \textbf{Recall} $\uparrow$
& \textbf{F1} $\uparrow$
& \textbf{Neg Acc} $\uparrow$
& \textbf{FP Rate} $\downarrow$
& \textbf{FN Rate} $\downarrow$
& \textbf{Unknown Rate} $\downarrow$
& \textbf{n} \\
\hline

\multirow{6}{*}{LLaVA}
& Normal
& 0.94 & 0.96 & 0.92 & 0.94 & 0.96 & 0.04 & 0.08 & 0.00 & 300 \\
& No Image
& 0.50 & 0.00 & 0.00 & 0.00 & 1.00 & 0.00 & 1.00 & 0.00 & 300 \\
& Black ($p=0.5$)
& 0.88 & 0.98 & 0.78 & 0.87 & 0.99 & 0.01 & 0.22 & 0.00 & 300 \\
& Black ($p=0.75$)
& 0.78 & 0.93 & 0.61 & 0.74 & 0.95 & 0.05 & 0.39 & 0.00 & 300 \\
& Noise ($p=0.5$)
& 0.91 & 0.94 & 0.88 & 0.91 & 0.94 & 0.06 & 0.12 & 0.00 & 300 \\
& Noise ($p=0.75$)
& 0.74 & 0.90 & 0.53 & 0.67 & 0.94 & 0.06 & 0.47 & 0.00 & 300 \\
\hline
\multirow{6}{*}{Gemma3}
& Normal
& 0.89 & 0.89 & 0.96 & 0.92 & 0.88 & 0.12 & 0.04 & 0.03 & 300 \\
& No Image
& 0.50 & 0.50 & 0.36 & 0.42 & 0.64 & 0.36 & 0.64 & 0.00 & 300 \\
& Black ($p=0.5$)
& 0.80 & 0.85 & 0.86 & 0.85 & 0.84 & 0.16 & 0.14 & 0.06 & 300 \\
& Black ($p=0.75$)
& 0.74 & 0.82 & 0.82 & 0.82 & 0.81 & 0.19 & 0.18 & 0.09 & 300 \\
& Noise ($p=0.5$)
& 0.51 & 0.77 & 0.98 & 0.86 & 0.49 & 0.51 & 0.02 & 0.36 & 300 \\
& Noise ($p=0.75$)
& 0.13 & 0.60 & 1.00 & 0.75 & 0.00 & 1.00 & 0.00 & 0.78 & 300 \\
\hline
\multirow{6}{*}{Qwen-32B}
& Normal
& 0.94 & 0.97 & 0.94 & 0.96 & 0.97 & 0.03 & 0.06 & 0.01 & 300 \\
& No Image
& 0.50 & 0.00 & 0.00 & 0.00 & 1.00 & 0.00 & 1.00 & 0.00 & 300 \\
& Black ($p=0.5$)
& 0.85 & 0.99 & 0.79 & 0.88 & 0.99 & 0.01 & 0.21 & 0.05 & 300 \\
& Black ($p=0.75$)
& 0.80 & 0.98 & 0.67 & 0.80 & 0.99 & 0.01 & 0.33 & 0.05 & 300 \\
& Noise ($p=0.5$)
& 0.83 & 0.92 & 0.88 & 0.90 & 0.92 & 0.08 & 0.12 & 0.08 & 300 \\
& Noise ($p=0.75$)
& 0.15 & 0.96 & 0.68 & 0.79 & 0.96 & 0.04 & 0.32 & 0.81 & 300 \\
\hline
\multirow{6}{*}{Qwen3-8B}
& Normal
& 0.97 & 0.99 & 0.95 & 0.97 & 0.99 & 0.01 & 0.05 & 0.00 & 300 \\
& No Image
& 0.50 & 0.00 & 0.00 & 0.00 & 1.00 & 0.00 & 1.00 & 0.00 & 300 \\
& Black ($p=0.5$)
& 0.86 & 0.98 & 0.74 & 0.84 & 0.99 & 0.01 & 0.26 & 0.00 & 300 \\
& Black ($p=0.75$)
& 0.77 & 0.93 & 0.58 & 0.71 & 0.96 & 0.04 & 0.42 & 0.00 & 300 \\
& Noise ($p=0.5$)
& 0.91 & 0.97 & 0.84 & 0.90 & 0.97 & 0.03 & 0.16 & 0.00 & 300 \\

& Noise ($p=0.75$)
& 0.70 & 0.86 & 0.46 & 0.60 & 0.93 & 0.07 & 0.54 & 0.00 & 300 \\
\hline
\multirow{6}{*}{Qwen3-4B}
& Normal
& 0.96 & 0.99 & 0.93 & 0.96 & 0.99 & 0.01 & 0.07 & 0.00 & 300 \\

& No Image
& 0.50 & 0.00 & 0.00 & 0.00 & 1.00 & 0.00 & 1.00 & 0.00 & 300 \\

& Black ($p=0.5$)
& 0.87 & 1.00 & 0.74 & 0.85 & 1.00 & 0.00 & 0.26 & 0.00 & 300 \\

& Black ($p=0.75$)
& 0.78 & 0.96 & 0.59 & 0.73 & 0.97 & 0.03 & 0.41 & 0.00 & 300 \\

& Noise ($p=0.5$)
& 0.89 & 0.95 & 0.83 & 0.88 & 0.96 & 0.04 & 0.17 & 0.00 & 300 \\

& Noise ($p=0.75$)
& 0.62 & 0.83 & 0.34 & 0.48 & 0.93 & 0.07 & 0.66 & 0.03 & 300 \\
\hline
\multirow{6}{*}{InternVL}
& Normal
& 0.98 & 0.98 & 0.97 & 0.98 & 0.98 & 0.02 & 0.03 & 0.00 & 300 \\

& No Image
& 0.47 & 0.00 & 0.00 & 0.00 & 1.00 & 0.00 & 1.00 & 0.11 & 300 \\

& Black ($p=0.5$)
& 0.86 & 0.98 & 0.74 & 0.85 & 0.99 & 0.01 & 0.26 & 0.00 & 300 \\

& Black ($p=0.75$)
& 0.81 & 0.97 & 0.64 & 0.77 & 0.98 & 0.02 & 0.36 & 0.00 & 300 \\

& Noise ($p=0.5$)
& 0.93 & 0.96 & 0.91 & 0.93 & 0.96 & 0.04 & 0.09 & 0.00 & 300 \\

& Noise ($p=0.75$)
& 0.72 & 0.80 & 0.61 & 0.69 & 0.85 & 0.16 & 0.39 & 0.01 & 300 \\
\hline
\multirow{6}{*}{Molmo}
& Normal
& 0.94 & 0.95 & 0.94 & 0.95 & 0.95 & 0.05 & 0.06 & 0.01 & 300 \\

& No Image
& 0.51 & 1.00 & 0.01 & 0.01 & 1.00 & 0.00 & 0.99 & 0.00 & 300 \\

& Black ($p=0.5$)
& 0.86 & 0.96 & 0.76 & 0.85 & 0.97 & 0.03 & 0.24 & 0.01 & 300 \\

& Black ($p=0.75$)
& 0.77 & 0.91 & 0.62 & 0.74 & 0.94 & 0.06 & 0.38 & 0.02 & 300 \\

& Noise ($p=0.5$)
& 0.91 & 0.94 & 0.89 & 0.91 & 0.94 & 0.06 & 0.11 & 0.01 & 300 \\

& Noise ($p=0.75$)
& 0.69 & 0.90 & 0.75 & 0.82 & 0.91 & 0.09 & 0.25 & 0.16 & 300 \\

\hline
\end{tabular}
\end{table*}

Table~\ref{tab:amber_gen_results} presents the full AMBER results across all models and visual conditions. Overall, the results show a pattern broadly consistent with the findings in the main paper. Under moderate global degradations such as \textit{Black} and \textit{Blur}, most models exhibit only limited or gradual performance changes relative to the original-image condition, especially when the degradation level is not too severe. By contrast, the \textit{No Image} condition often leads to much larger deterioration, indicating that completely removing visual input remains substantially more disruptive than weakening fine-grained visual evidence.

At the metric level, \textit{Cover} generally decreases under stronger perturbations, while \textit{CHAIR} and \textit{Hal} often increase, suggesting a reduction in grounded content coverage together with a higher tendency toward hallucinated generation. The overall pattern therefore supports the main observation of this work: partial degradation of visual evidence does not always cause catastrophic behavioral collapse, but complete removal of the image typically does.

We also note that model behavior on AMBER is more heterogeneous than on closed-form benchmarks. In particular, some models show extreme outputs under the \textit{No Image} condition, which likely reflects model-specific generation behavior or evaluator interactions in open-ended settings. For this reason, we treat AMBER primarily as complementary evidence rather than the main basis for our controlled analysis.

\subsection{Detailed Results with Expanded Answer Options}
\label{app:unknown_detailed_results}

\begin{table*}[t]
\centering
\small
\setlength{\tabcolsep}{7pt}
\renewcommand{\arraystretch}{0.85}
\caption{\textbf{Decision-level support for affirmative predictions under visual interventions} (ground truth = \textit{Yes}). We report mean and median logit margins, mean and median affirmative probabilities ($p_{\text{yes}}$), and confidence-shift statistics relative to the \textit{Original} condition. Here, $\mathbb{E}[\delta]$ denotes the expected reduction in affirmative support.}
\label{tab:prob_analysis}
\begin{tabular}{l | cc | cc | ccc}
\toprule
\multirow{2}{*}{\textbf{Condition}} & \multicolumn{2}{c|}{\textbf{Margin} $\uparrow$} & \multicolumn{2}{c|}{\textbf{Probability ($p_{\text{yes}}$)} $\uparrow$} & \multicolumn{3}{c}{\textbf{Confidence Shifts}} \\
\cmidrule(lr){2-3} \cmidrule(lr){4-5} \cmidrule(lr){6-8}
& Mean & Median & Mean & Median & $\mathbb{E}[\delta] \uparrow$ & $\Pr(\delta>0) \uparrow$ & $\Pr(\delta>1) \uparrow$ \\
\midrule
Original    & 4.56 & 5.12 & 0.93 & 0.99 & --     & --     & --     \\
Black Mask  & 2.20 & 2.44 & 0.77 & 0.92 & 2.36 & 0.86 & 0.62 \\
Black Box   & 1.14 & 0.81 & 0.61 & 0.69 & 3.42 & 0.88 & 0.67 \\
No Image    & -0.14 & 0.00 & 0.47 & 0.50 & 4.72 & 0.92 & 0.89 \\
Entity Swap & 1.05 & 0.94 & 0.60 & 0.72 & 3.53 & 0.87 & 0.68 \\
\bottomrule
\end{tabular}
\end{table*}

In the main paper, for the setting with expanded answer options, we report only the \textit{Unknown Rate} as the primary quantity of interest. Here, we provide the full results for completeness, including accuracy, precision, recall, F1, negative accuracy, false-positive rate, and false-negative rate.

Table~\ref{tab:unknown_detailed_results} shows the detailed results across all models and visual conditions on POPE when the model is allowed to answer \textit{Yes}, \textit{No}, or \textit{Unknown}. Overall, the results suggest that introducing an \textit{Unknown} option does not fundamentally change the main behavioral pattern observed in the paper. For most models, performance under moderate visual degradation remains relatively stable, whereas the \textit{No Image} condition leads to a much larger drop.

The detailed metrics also help clarify the role of the \textit{Unknown} option. For several models, the increase in \textit{Unknown Rate} under stronger perturbations is accompanied by changes in both recall and false-negative rate, indicating that the model becomes more likely to abstain or avoid committing to positive predictions when visual evidence is severely weakened. At the same time, this behavior is not uniform across models: some models rarely choose \textit{Unknown}, while others use it much more aggressively under severe degradation.

These results therefore support the observation in the main text that the availability of an explicit \textit{Unknown} option does not by itself resolve the benchmark--grounding mismatch. Instead, it mainly reveals model-specific differences in uncertainty expression under degraded visual input.

\subsection{Detailed Statistics for Decision-Level Support}
\label{app:prob_analysis}

In the main paper, we visualize decision-level support for affirmative predictions using box plots. To complement that distributional view, we report here the corresponding summary statistics on the POPE positive subset ($n=149$, ground truth = \textit{Yes}).

Table~\ref{tab:prob_analysis} summarizes the model's support for the affirmative answer under different visual interventions in terms of both logit margin and affirmative probability $p_{\text{yes}}$. In addition to the mean and median values, we also report confidence-shift statistics relative to the \textit{Original} condition. Specifically, $\mathbb{E}[\delta]$ measures the expected reduction in affirmative support, while $\Pr(\delta > 0)$ and $\Pr(\delta > 1)$ quantify how often the intervention lowers support, and how often the drop exceeds a larger threshold.

The results are consistent with the distributional patterns shown in the main text. Relative to the \textit{Original} condition, all interventions reduce affirmative support on the positive subset. The effect is strongest under \textit{No Image}, where both the mean margin and mean affirmative probability fall close to the decision boundary, indicating a near-complete loss of positive support. Localized interventions such as \textit{Black Mask}, \textit{Black Box}, and \textit{Entity Swap} also substantially reduce support, even though they do not remove the entire image.

These results further support our main claim that benchmark predictions can remain behaviorally stable even when the internal support for the correct affirmative answer has already weakened considerably. In other words, decision correctness alone may understate the extent to which fine-grained visual evidence has been lost or degraded.

\begin{figure*}[htbp]
    \centering
    \includegraphics[width=0.95\textwidth]{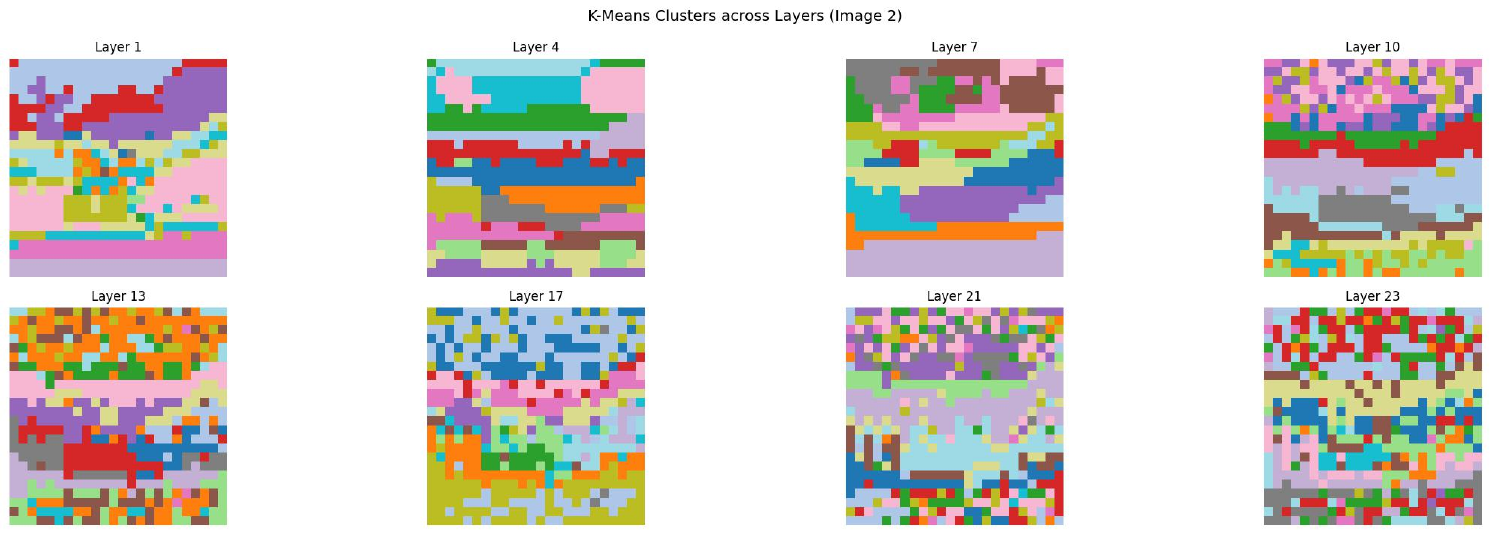}
    \caption{Qualitative visualization of region-level clustering structures and spatial token evolution.}
    \label{fig:cluster_case}
\end{figure*}

\subsection{Qualitative Case Study of Spatial Representation Evolution}
\label{app:cluster_case}

To provide an intuitive illustration of the representational trends discussed in the main paper, we present a qualitative case study of spatial token clustering across layers.

Figure~\ref{fig:cluster_case} visualizes the $k$-means clustering assignments of spatial tokens at different layers of the visual encoder. Each colored cell corresponds to one spatial token, and colors indicate cluster membership ($k=16$). The spatial layout of tokens follows the original patch grid of the image.

In early layers (e.g., Layers 1 and 4), the clusters exhibit relatively coherent spatial regions, with contiguous patches often belonging to the same cluster. This suggests that local visual structure and spatial organization are still preserved in the representation.

As depth increases, however, the clustering structure becomes progressively more fragmented. In later layers (e.g., Layers 17, 21, and 23), cluster assignments appear increasingly mixed and spatially irregular. Neighboring patches are less likely to share the same cluster, indicating that the representations of different spatial regions become more intertwined.

This qualitative pattern aligns with the quantitative results reported in the main paper, where measures such as inter-block similarity and effective rank indicate a gradual loss of spatial distinctiveness across layers. The visualization therefore provides an intuitive illustration of how spatial token structure evolves during visual encoding.

\end{document}